\documentclass[11pt,a4paper]{article}
\usepackage[hyperref]{naaclhlt2018}
\usepackage{times}
\usepackage{latexsym}
\usepackage{graphicx}
\usepackage{tabularx}
\usepackage{wrapfig}
\usepackage[normalem]{ulem}
\usepackage{amsmath}
\usepackage{subcaption}
\usepackage{xcolor}
\usepackage{multirow}
\usepackage{multicol}
\usepackage{makecell}
\usepackage{amssymb}
\def\squiggly{\bgroup \markoverwith{\textcolor{red}{\lower3.5\p@\hbox{\sixly \char58}}}\ULon}

\usepackage{url}
\usepackage{xcolor} 

\makeatletter
\newcommand{\thickhline}{%
    \noalign {\ifnum 0=`}\fi \hrule height 1pt
    \futurelet \reserved@a \@xhline
}
\makeatother

\usepackage{url}

\aclfinalcopy 

\title{ThisIsCompetition at SemEval-2019 Task 9:\\BERT is unstable for out-of-domain samples}

\author{Cheoneum Park*$^1$ \and
  Juae Kim*$^2$ \and
  Hyeon-gu Lee*$^1$ \and
  Reinald Kim Amplayo*$^3$ \\
  \textbf{Harksoo Kim}$^1$ \and
  \textbf{Jungyun Seo}$^2$ \and 
  \textbf{Changki Lee}$^1$ \\
  \textbf{(* equal contribution)} \\
  $^1$Kangwon National University, South Korea \\
  $^2$Sogang University, South Korea \\
  $^3$University of Edinburgh, UK
}

\date{}

\begin{document}
\maketitle
\begin{abstract}
  This paper describes our system, Joint Encoders for Stable Suggestion Inference (\textbf{JESSI}), for the SemEval 2019 Task 9: Suggestion Mining from Online Reviews and Forums. JESSI is a combination of two sentence encoders: (a) one using multiple pre-trained word embeddings learned from log-bilinear regression (GloVe) and translation (CoVe) models, and (b) one on top of word encodings from a pre-trained deep bidirectional transformer (BERT). We include a domain adversarial training module when training for out-of-domain samples. Our experiments show that while BERT performs exceptionally well for in-domain samples, several runs of the model show that it is unstable for out-of-domain samples. The problem is mitigated tremendously by (1) combining BERT with a non-BERT encoder, and (2) using an RNN-based classifier on top of BERT. Our final models obtained second place with 77.78\% F-Score on Subtask A (i.e. in-domain) and achieved an F-Score of 79.59\% on Subtask B (i.e. out-of-domain), even without using any additional external data.
\end{abstract}

\section{Introduction}

Opinion mining \cite{pang2007opinion} is a huge field that covers many NLP tasks ranging from sentiment analysis \cite{liu2012sentiment}, aspect extraction \cite{mukherjee2012aspect}, and opinion summarization \cite{ku2006opinion}, among others. Despite the vast literature on opinion mining, the task on suggestion mining has given little attention. Suggestion mining \cite{brun2013suggestion} is the task of collecting and categorizing suggestions about a certain product. This is important because while opinions indirectly give hints on how to improve a product (e.g. analyzing reviews), suggestions are direct improvement requests (e.g. tips, advice, recommendations) from people who have used the product.

To this end, \citet{negi2019semeval} organized a shared task specifically on suggestion mining called SemEval 2019 Task 9: Suggestion Mining from Online Reviews and Forums. The shared task is composed of two subtasks, Subtask A and B. In Subtask A, systems are tasked to predict whether a sentence of a certain domain (i.e. electronics) entails a suggestion or not given a training data of the same domain. In Subtask B, systems are tasked to do suggestion prediction of a sentence from another domain (i.e. hotels). Organizers observed four main challenges: (a) sparse occurrences of suggestions; (b) figurative expressions; (c) different domains; and (d) complex sentences. While previous attempts \cite{ramanand2010wishful,brun2013suggestion,negi2015towards} made use of human-engineered features to solve this problem, the goal of the shared task is to leverage the advancements seen on neural networks, by providing a larger dataset to be used on data-intensive models to achieve better performance.

This paper describes our system \textbf{JESSI} (Joint Encoders for Stable Suggestion Inference). JESSI is built as a combination of two neural-based encoders using multiple pre-trained word embeddings, including BERT \cite{devlin2018bert}, a pre-trained deep bidirectional transformer that is recently reported to perform exceptionally well across several tasks. The main intuition behind JESSI comes from our finding that although BERT gives exceptional performance gains when applied to in-domain samples, it becomes unstable when applied to out-of-domain samples, even when using a domain adversarial training \cite{ganin2016domain} module. This problem is mitigated using two tricks: (1) jointly training BERT with a CNN-based encoder, and (2) using an RNN-based encoder on top of BERT before feeding to the classifier.

JESSI is trained using only the datasets given on the shared task, without using any additional external data. Despite this, JESSI performs second on Subtask A with an F1 score of 77.78\% among 33 other team submissions. It also performs well on Subtask B with an F1 score of 79.59\%.

\section{Related Work}

\paragraph{Suggestion Mining}

The task of detecting suggestions in sentences is a relatively new task, first mentioned in \citet{ramanand2010wishful} and formally defined in \citet{negi2015towards}. Early systems used manually engineered patterns \cite{ramanand2010wishful} and rules \cite{brun2013suggestion}, and linguistically motivated features \cite{negi2015towards} trained on a supervised classifier \cite{negi2016study}. Automatic mining of suggestion has also been suggested \cite{dong2013automated}. Despite the recent successes of neural-based models, only few attempts were done, by using neural network classifiers such as CNNs and LSTMs \cite{negi2016study}, by using part-of-speech embeddings to induce distant supervision \cite{negi2017inducing}. Since neural networks are data hungry models, a large dataset is necessary to optimize the parameters. SemEval 2019 Task 9 \cite{negi2019semeval} enables training of deeper neural models by providing a much larger training dataset.

\paragraph{Domain Adaptation}

In text classification, training and test data distributions can be different, and thus domain adaptation techniques are used. These include non-neural methods that map the semantics between domains by aligning the vocabulary \cite{basili2009cross,pan2010cross} and generating labeled samples \cite{wan2009co,yu2016learning}. Neural methods include the use of stacked denoising autoencoders \cite{glorot2011domain}, variational autoencoders \cite{saito2017asymmetric,ruder2018strong}. Our model uses a domain adversarial training module \cite{ganin2016domain}, an elegant way to effectively transfer knowledge between domains by training a separate domain classifier using an adversarial objective.

\paragraph{Language Model Pretraining}

Inspired from the computer vision field, where ImageNet \cite{deng2009imagenet} is used to pretrain models for other tasks \cite{huh2016what}, many recent attempts in the NLP community are successful on using language modeling as a pretraining step to extract feature representations \cite{peters2018deep}, and to fine-tune NLP models \cite{radford2018improving,devlin2018bert}. BERT \cite{devlin2018bert} is the most recent inclusion to these models, where it uses a deep bidirectional transformer trained on masked language modeling and next sentence prediction objectives. \citet{devlin2018bert} reported that BERT shows significant increase in improvements on many NLP tasks, and subsequent studies have shown that BERT is also effective on harder tasks such as open-domain question answering \cite{yang2019end}, multiple relation extraction \cite{wang2019extracting}, and table question answering \cite{hwang2019comprehensive}, among others. In this paper, we also use BERT as an encoder, show its problem on out-of-domain samples, and mitigate the problem using multiple tricks.

\section{Joint Encoders for Stable Suggestion Inference}

\begin{figure}
    \centering
    \includegraphics[width=0.47\textwidth]{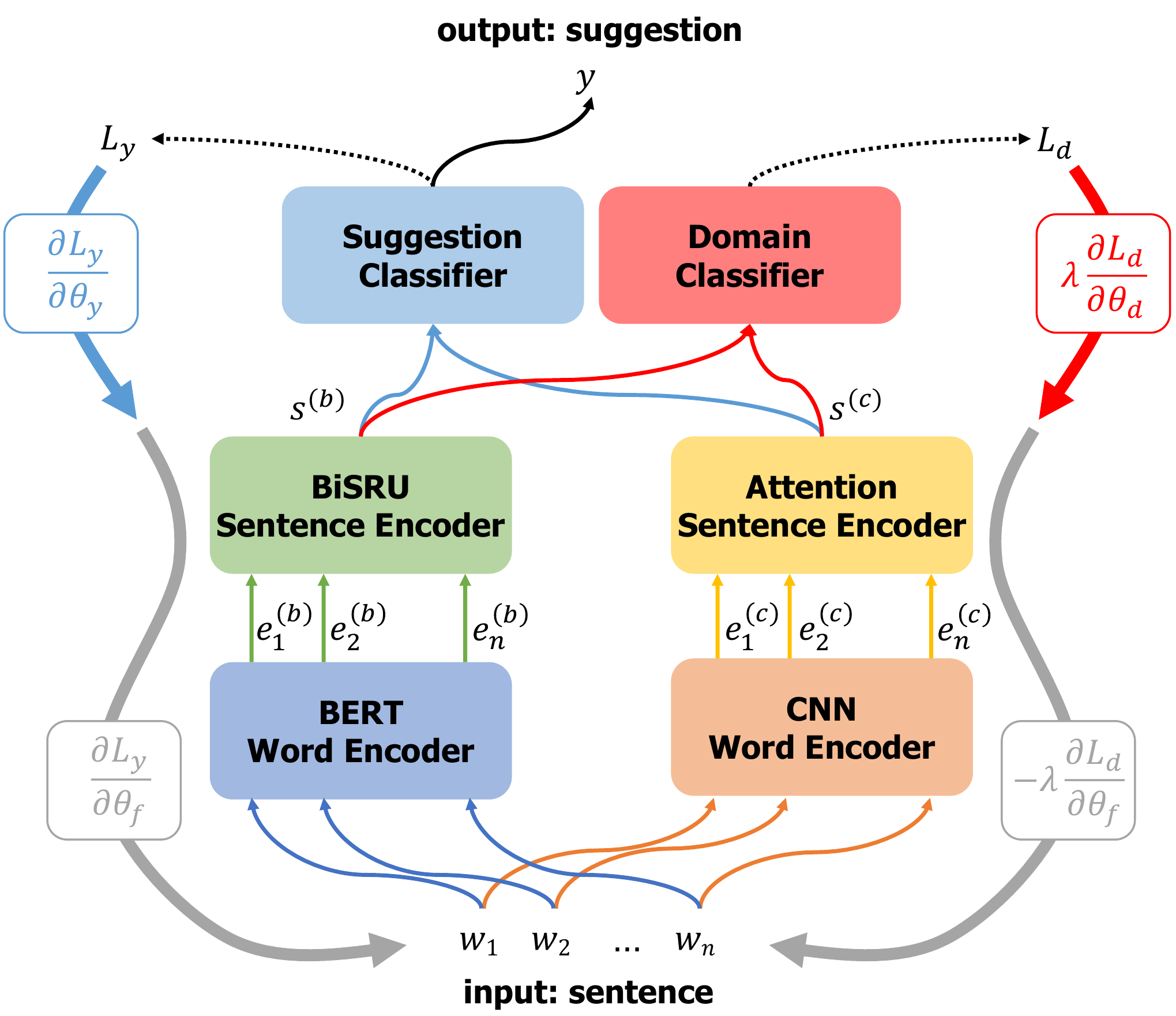}
    \caption{The overall architecture of JESSI for Subtask B. The thinner arrows correspond to the forward propagations, while the thicker arrows correspond to the backward propagations, where gradient calculations are indicated. For Subtask A, a CNN encoder is used instead of the BiSRU encoder, and the domain adversarial training module is not used.}
    \label{fig:model}
\end{figure}

We present our model \textbf{JESSI}, which stands for Joint Encoders for Stable Suggestion Inference, shown in Figure \ref{fig:model}. Given a sentence $x=\{w_1,w_2,...,w_n\}$, JESSI returns a binary suggestion label $y=\{0,1\}$. JESSI consists of four important components: (1) A BERT-based encoder that leverages general knowledge acquired from a large pre-trained language model, (2) A CNN-based encoder that learns task-specific sentence representations, (3) an MLP classifier that predicts the label given the joint encodings, and (4) a domain adversarial training module that prevents the model to distinguish between the two domains.

\paragraph{BERT-based Encoder}

Fine-tuning a pre-trained BERT \cite{devlin2018bert} classifier then using the separately produced classification encoding \textsc{[CLS]} has shown to produce significant improvements. Differently, JESSI uses a pre-trained BERT as a word encoder, that is instead of using \textsc{[CLS]}, we use the word encodings $e^{(b)}_1,e^{(b)}_2,...,e^{(b)}_n$ produced by BERT. BERT is still fine-tuned during training.

We append a sentence encoder on top of BERT, that returns a sentence representation $s^{(b)}$, which is different per subtask. For Subtask A, we use a CNN encoder with max pooling \cite{kim2014convolutional} to create the sentence embedding. For Subtask B, we use a bidirectional simple recurrent units \cite[BiSRU]{lei2018simple}, a type of RNN that is highly parallelizable, as the sentence encoder.

\paragraph{CNN-based Encoder}

To make the final classifier more task-specific, we use a CNN-based encoder that is trained from scratch. Specifically, we employ a concatenation of both pre-trained GloVe \cite{pennington2014glove} and CoVe \cite{mccann2017learned} word embeddings as input $w_i, 1 \leq i \leq n$. Then, we do convolution operations $\text{Conv}(w_i, h_j)$ using multiple filter sizes $h_j$ to a window of $h_j$ words. We use different paddings for different filter sizes such that the number of output for each convolution operation is $n$. Finally, we concatenate the outputs to obtain the word encodings, i.e. $e^{(c)}_i = \oplus_j (\text{Conv}(w_i, h_j))$, where $\oplus$ is the sequence concatenate operation.

We pool the word encodings using attention mechanism to create a sentence representation $s^{(c)}$. That is, we calculate attention weights using a latent variable $v$ that measures the importance of the words $e^{(c)}_i$, i.e., $a_i = \text{softmax}(v^\top f(e^{(c)}_i))$, where $f(\cdot)$ is a nonlinear function. We then use $a_i$ to weight-sum the words into one encoding, i.e., $s^{(c)} = \sum_i a_i e^{(c)}_i$.

\paragraph{Suggestion Classifier}

Finally, we use a multi-layer perceptron (MLP) as our classifier, using a concatenation of outputs from both the BERT- and CNN-based encoders, i.e., $p(y) = \text{MLP}_y([s^{(b)};s^{(c)}])$. Training is done by minimizing the cross entropy loss, i.e., $\mathbb{L} = - \log{p(y)}$.

\paragraph{Domain Adversarial Training}

For Subtask B, the model needs to be able to classify out-of-domain samples. Using the model as is decreases performance significantly because of cross-domain differences. To this end, we use a domain adversarial training module \cite{ganin2016domain} to prevent the classifier on distinguishing differences between domains. Specifically, we create another MLP classifier that classifies the \textit{domain} of the text using the concatenated sentence encoding with \textit{reverse gradient function} $\text{GradRev}(\cdot)$, i.e., $p(d) = \text{MLP}_d(\text{GradRev}([s^{(b)};s^{(c)}]))$. The reverse gradient function is a function that performs equivalently with the identity function when propagating forward, but reverses the sign of the gradient when propagating backward. 

Through this, we eliminate the possible ability of the classifier to distinguish the domains of the text.
We train the domain classifier using the available trial datasets for each domain. We also use a cross entropy loss as the objective of this classifier. Overall, the objective of JESSI is to minimize the following loss: $\mathbb{L} = - \log{p(y)} - \lambda \log{p(d)}$, where $\lambda$ is set increasingly after each epoch, following \citet{ganin2016domain}.

\section{Experimental Setup}

\paragraph{Dataset and Preprocessing}

We use the dataset provided in the shared task: a training dataset from the electronics domain, and labeled trial and unlabeled test datasets from both the electronics and hotels domain. Table \ref{tab:dataset} summarizes the dataset statistics and shows the distribution differences between two domains. During training, we use the labeled training dataset to train the suggestions classifier, and trial datasets, without the suggestion labels, to train the domains classifier. For preprocessing, we lowercased and tokenized using the Stanford CoreNLP toolkit\footnote{\url{https://stanfordnlp.github.io/CoreNLP/}} \cite{manning2014stanford}.

\begin{table}[t]
    \centering
    \begin{tabular}{l|cc}
        \thickhline
        Subtask & A & B \\
        \thickhline
        Domain & Electronics & Hotels \\
        \#Training & 8,230 & 0 \\
        \#Trial & 592 & 808 \\
        \#Test & 833 & 824 \\
        \#Vocabulary & 10,897 & 3,570 \\
        Ave. Tokens & 19.0 & 16.8 \\
        \thickhline
    \end{tabular}
    \caption{Dataset Statistics}
    \label{tab:dataset}
\end{table}

\paragraph{Implementation}

We use the pre-trained BERT models\footnote{\url{https://github.com/google-research/bert}} provided by the original authors to initialize the parameters of BERT. We use BERT-large for Subtask A and BERT-base for Subtask B\footnote{Due to memory limitations, we are limited to use the smaller BERT model for Subtask B. We expect an increase in performance when BERT-large is used.}. For our CNNs, we use three filters with sizes $\{3,5,7\}$, each with 200 dimensions. For the BiSRU, we use hidden states with 150 dimensions and stack with two layers. The MLP classifier contains two hidden layers with 300 dimensions.

We use dropout \cite{srivastava2014dropout} on all nonlinear connections with a dropout rate of 0.5. We also use an $l_2$ constraint of 3. During training, we use mini-batch size of 32. Training is done via stochastic gradient descent over shuffled mini-batches with the Adadelta \cite{zeiler2012adadelta} update rule. We perform early stopping using the trial sets. Moreover, since the training set is relatively small, multiple runs lead to different results. To handle this, we perform an ensembling method as follows. We first run 10-fold validation over the training data, resulting into ten different models. We then pick the top three models with the highest performances, and pick the class with the most model predictions.

\section{Experiments}
\label{sec:experiments}

In this section, we show our results and experiments. We denote \textsc{JESSI-A} as our model for Subtask A (i.e., \textsc{BERT$\rightarrow$CNN+CNN$\rightarrow$Att}), and \textsc{JESSI-B} as our model for Subtask B (i.e., \textsc{BERT$\rightarrow$BiSRU+CNN$\rightarrow$Att+DomAdv}). The performance of the models is measured and compared using the F1-score.

\paragraph{Ablation Studies}

\begin{table}[t]
    \centering
    \begin{subtable}{0.5\textwidth}
        \centering
        \begin{tabular}{lr}
            \thickhline
            Model & F-Score \\
            \thickhline
            \textsc{JESSI-A} & 88.78 \\
            { }{ } + \textsc{BERT$\rightarrow$BiSRU} & 86.01 \\
            { }{ } -- \textsc{CNN$\rightarrow$Att} & 85.14 \\
            { }{ } -- \textsc{BERT$\rightarrow$CNN} & 83.89 \\
            \thickhline
        \end{tabular}
        \caption{Subtask A}
        \label{tab:ablation_a}
    \end{subtable}
    \begin{subtable}{0.5\textwidth}
        \centering
        \begin{tabular}{lr}
            \thickhline
            Model & F-Score \\
            \thickhline
            \textsc{JESSI-B} & 87.31 \\
            { }{ } -- \textsc{CNN$\rightarrow$Att} & 84.01 \\
            { }{ } -- \textsc{BERT$\rightarrow$BiSRU} & 81.13 \\
            { }{ } + \textsc{BERT$\rightarrow$CNN} & 70.21 \\
            { }{ } -- \textsc{DomAdv} & 47.48 \\
            \thickhline
        \end{tabular}
        \caption{Subtask B}
        \label{tab:ablation_b}
    \end{subtable}
    \caption{Ablation results for both subtasks using the provided trial sets. The + denotes a \textit{replacement} of the BERT-based encoder, while the -- denotes a \textit{removal} of a specific component.}
    \label{tab:ablation}
\end{table}

We present in Table \ref{tab:ablation} ablations on our models. Specifically, we compare JESSI-A with the same model, but without the CNN-based encoder, without the BERT-based encoder, and with the CNN sentence encoder of the BERT-based encoder replaced with the BiSRU variant. We also compare JESSI-B with the same model, but without the CNN-based encoder, without the BERT-based encoder, without the domain adversarial training module, and with the BiSRU sentence encoder of the BERT-based encoder replaced with the CNN variant. The ablation studies show several observations. First, jointly combining both BERT- and CNN-based encoders help improve the performance on both subtasks. Second, the more effective sentence encoder for the BERT-based encoder (i.e., CNN versus BiSRU) differs for each subtask; the CNN variant is better for Subtask A, while the BiSRU variant is better for Subtask B. Finally, the domain adversarial training module is very crucial in achieving a significant increase in performance.

\paragraph{Out-of-Domain Performance}

During our experiments, we noticed that BERT is unstable when predicting out-of-domain samples, even when using the domain adversarial training module. We show in Table \ref{tab:out_of_domain} the summary statistics of the F-Scores of 10 runs of the following models: (a) vanilla \textsc{BERT} that uses the \textsc{[CLS]} classification encoding, (b-c) our BERT-based encoders \textsc{BERT$\rightarrow$CNN} and \textsc{BERT$\rightarrow$BiSRU} that use BERT as a word encoder and use an additional CNN/BiSRU as a sentence encoder, (d) \textsc{JESSI-B} that uses \textsc{BERT$\rightarrow$BiSRU} and \textsc{CNN$\rightarrow$Att} as joint encoders, and (e) \textsc{CNN$\rightarrow$Att} that does not employ BERT in any way. The results show that while \textsc{CNN$\rightarrow$Att} performs similarly on different runs, \textsc{BERT} performs very unstably, achieving varying F-Scores as low as zero and as high as 70.59, with a standard deviation of 31. Appending a CNN-based sentence encoder (i.e., \textsc{BERT$\rightarrow$CNN}) increases the performance, but worsens the stability of the model. Appending an RNN-based sentence encoder (i.e., \textsc{BERT$\rightarrow$BiSRU}) both increases the performance and improves the model stability. Finally, combining a separate CNN-based encoder (i.e., \textsc{JESSI-B}) improves the performance and stability further.

\begin{table}[t]
    \centering
    \begin{tabular}{@{}lrrrr@{}}
        \thickhline
        Model & min & max & mean & std \\
        \thickhline
        \textsc{BERT} & 0.00 & 70.59 & 22.52 & 31.0 \\
        \textsc{BERT$\rightarrow$CNN} & 0.00 & 74.62 & 28.23 & 34.1 \\
        \textsc{BERT$\rightarrow$BiSRU} & 54.00 & 88.83 & 74.86 & 8.8 \\
        \textsc{JESSI-B} & 69.28 & 89.21 & 82.41 & 5.6 \\
        \hline
        \textsc{CNN$\rightarrow$Att} & 68.19 & 77.06 & 72.50 & 2.5 \\
        \thickhline
    \end{tabular}
    \caption{Summary statistics of the F-Scores of 10 runs of different models on the trial set of Subtask B when doing a 10-fold validation over the available training data. All models include the domain adversarial training module (+\textsc{DomAdv}), which is omitted for brevity.}
    \label{tab:out_of_domain}
\end{table}

\paragraph{Test Set Results}

Table \ref{tab:final} presents how JESSI compared to the top performing models during the competition proper. Overall, JESSI-A ranks second out of 33 official submissions with an F-Score of 77.78\%. Although we were not able to submit JESSI-B during the submission phase, JESSI-B achieves an F-Score of 79.59\% on the official test set. This performance is similar to the performance of the model that obtained sixth place in the competition. We emphasize that JESSI does not use any labeled and external data for Subtask B, and thus is just exposed to the hotels domain using the available \textit{unlabeled} trial dataset, containing 808 data instances. We expect the model to perform better when additional data from the hotels domain.

\begin{table}[t]
    \centering
    \begin{subtable}{\columnwidth}
        \centering
        \begin{tabular}{lp{10em}r}
            \thickhline
            Rank & Model & F-Score \\
            \thickhline
            1 & OleNet & 78.12 \\
            2 & \textsc{JESSI-A} & 77.78 \\
            3 & m\_y & 77.61 \\
            \thickhline
        \end{tabular}
        \caption{Subtask A}
        \label{tab:final_a}
    \end{subtable}
    \begin{subtable}{\columnwidth}
        \centering
        \begin{tabular}{lp{10em}r}
            \thickhline
            Rank & Model & F-Score \\
            \thickhline
            1 & NTUA-ISLab & 85.80 \\
            2 & OleNet & 85.79 \\
            3 & NL-FIIT & 83.13 \\
            \hline
            * & \textsc{JESSI-B} & 79.59 \\
            11 & \textsc{CNN$\rightarrow$Att+DomAdv} & 64.86 \\
            \thickhline
        \end{tabular}
        \caption{Subtask B}
        \label{tab:final_b}
    \end{subtable}
    \caption{F-Scores of JESSI and top three models for each subtask. Due to time constraints, we were not able to submit JESSI-B during the competition. For clarity, we also show our final official submission (\textsc{CNN$\rightarrow$Att+DomAdv}).}
    \label{tab:final}
\end{table}

\paragraph{Performance by Length}

\begin{figure}[!t]
    \centering
    \begin{subfigure}[t]{0.47\textwidth}
        \includegraphics[width=\textwidth]{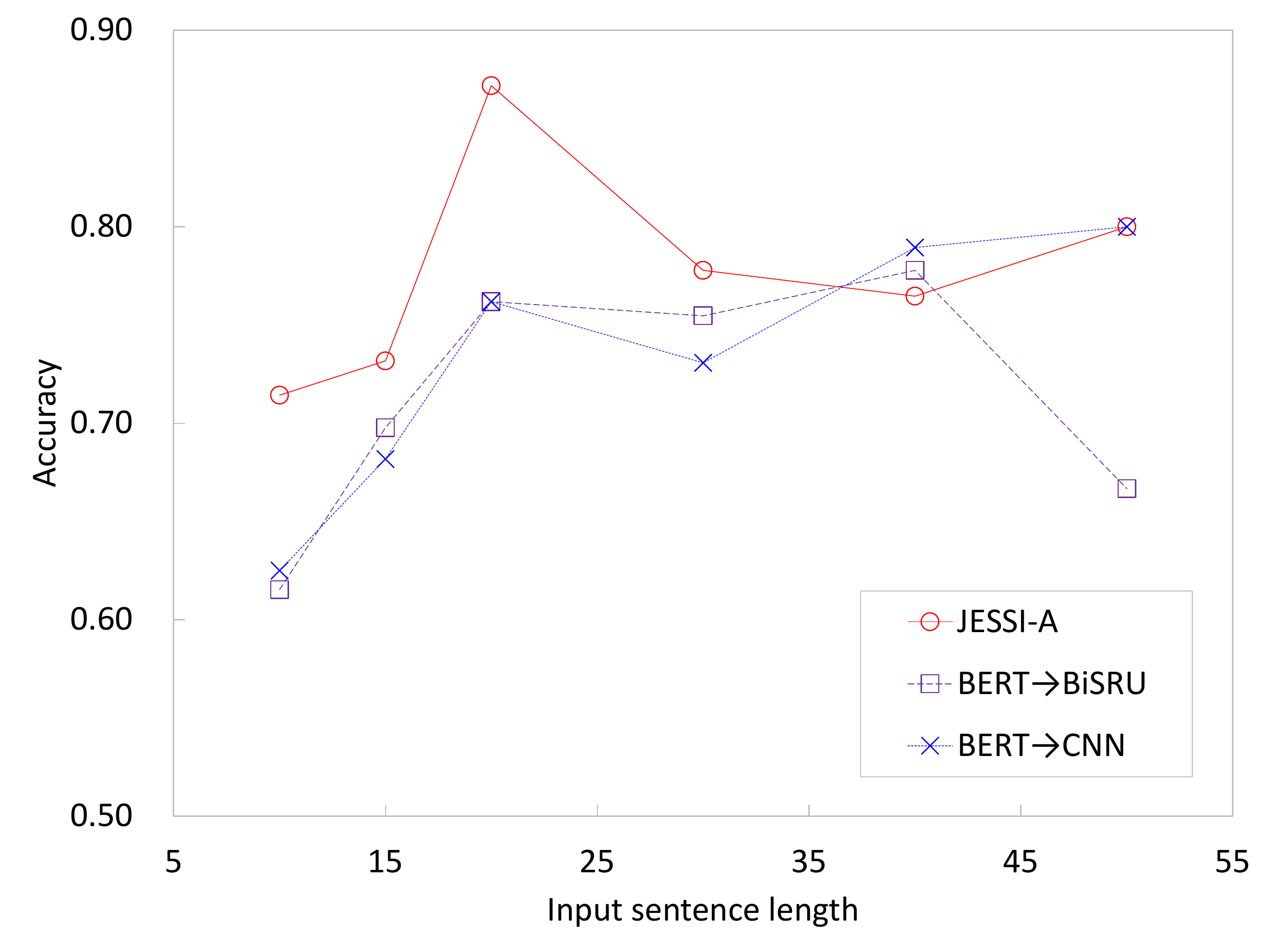}
        \caption{Subtask A}
        \label{subtask_a}
    \end{subfigure}
    \begin{subfigure}[t]{0.47\textwidth}
        \includegraphics[width=\textwidth]{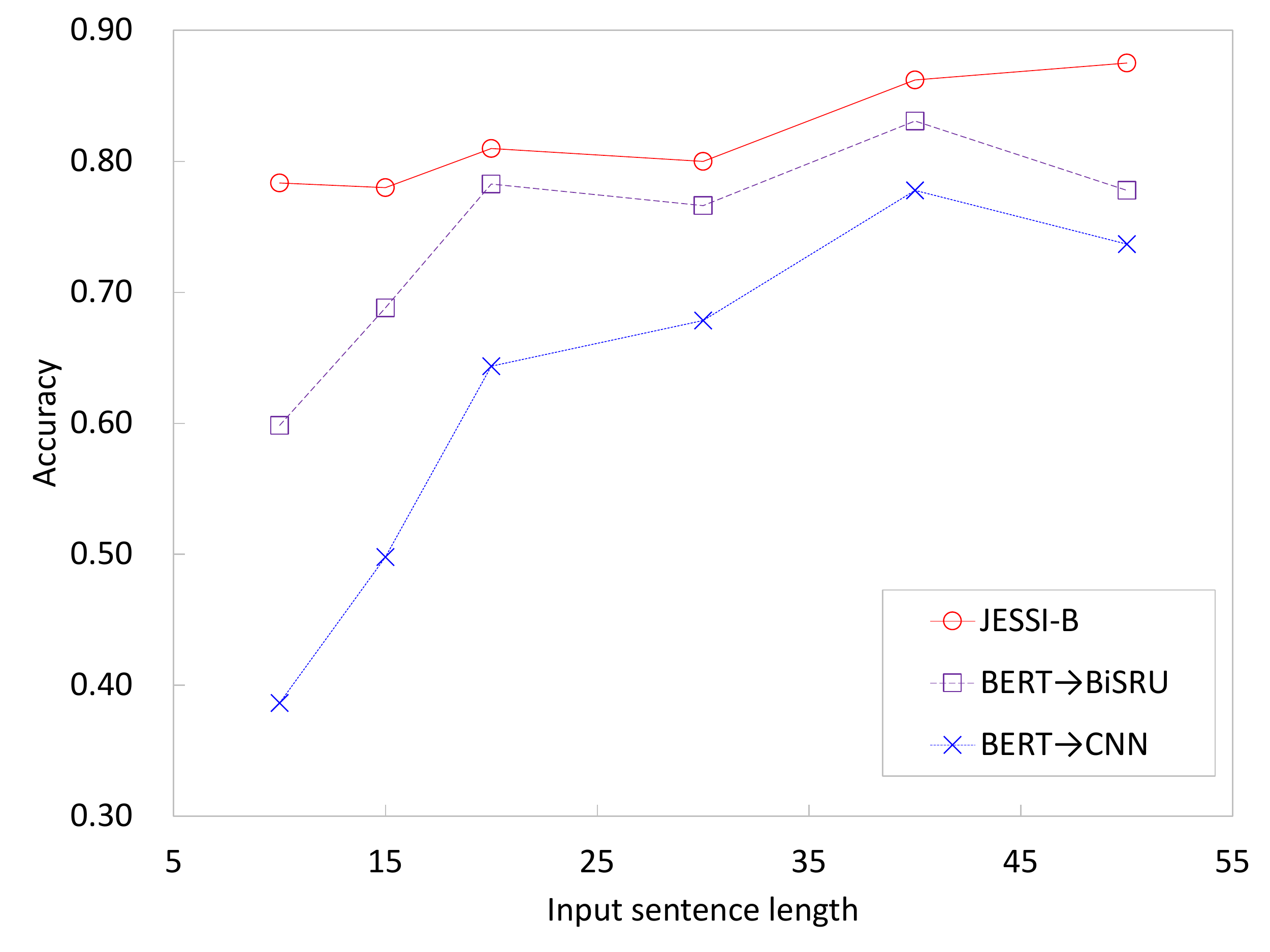}
        \caption{Subtask B}
        \label{subtask_b}
    \end{subfigure}
    \caption{Accuracy over various input sentence length on the test set.}
    \label{various_sent_acc}
\end{figure}

We compare the performance of models on data with varying lengths to further investigate the increase in performance of JESSI over other models.
More specifically, for each range of sentence length (e.g., from 10 to 20),
we look at the accuracy of JESSI-A, \textsc{BERT$\rightarrow$BiSRU}, and \textsc{BERT$\rightarrow$CNN} on Subtask A, and the accuracy of JESSI-B, \textsc{BERT$\rightarrow$BiSRU}, and \textsc{BERT$\rightarrow$CNN}, all with domain adversarial training module, on Subtask B.
Figure \ref{various_sent_acc} shows the plots of the experiments on both subtasks. On both experiments, JESSI outperforms the other models when the sentence length is short, suggesting that the increase in performance of JESSI can be attributed to its performance in short sentences. This is more evident in Subtask B, where the difference of accuracy between JESSI and the next best model is approximately 20\%.
We can also see a consistent increase in performance of JESSI over other models on Subtask B, which shows the robustness of JESSI when predicting out-of-domain samples.

\section{Conclusion}

We presented JESSI (Joint Encoders for Stable Suggestion Inference), our system for the SemEval 2019 Task 9: Suggestion Mining from Online Reviews and Forums. JESSI builds upon jointly combined encoders, borrowing pre-trained knowledge from a language model BERT and a translation model CoVe. We found that BERT alone performs bad and unstably when tested on out-of-domain samples. We mitigate the problem by appending an RNN-based sentence encoder above BERT, and jointly combining a CNN-based encoder. Results from the shared task show that JESSI performs competitively among participating models, obtaining second place on Subtask A with an F-Score of 77.78\%. It also performs well on Subtask B, with an F-Score of 79.59\%, even without using any additional external data.

\section*{Acknowledgement}
This research was supported by the MSIT (Ministry of Science ICT), Korea, under (National Program for Excellence in SW) (2015-0-00910) and (Artificial Intelligence Contact Center Solution) (2018-0-00605) supervised by the IITP(Institute for Information \& Communications Technology Planning \& Evaluation)

\bibliography{semeval2018}
\bibliographystyle{acl_natbib}

\end{document}